\documentclass{isprs} % isprs class modified 23-04-2019 (Dennis Wittich)
\usepackage{subfigure}
\usepackage{setspace}
\usepackage{geometry} % added 27-02-2014 Markus Englich
\usepackage{epstopdf}
\usepackage[labelsep=period]{caption}  % added 14-04-2016 Markus Englich - Recommendation by Sebastian Brocks
\usepackage[british]{babel} 
\usepackage[hang]{footmisc}
\usepackage{amsmath}
\usepackage[dvipsnames]{xcolor}
\usepackage{colortbl}

\begin{document}

\title{In-Field 3D Wheat Head Instance Segmentation From TLS Point Clouds Using Deep Learning Without Manual Labels}
\date{}

% KAO: Remove extra spacing

% Anonymous submissions, authors' names should not be visible
% \author{
%  Orhan Altan\textsuperscript{1}, Ian Dowman\textsuperscript{2}, Florent Lafarge\textsuperscript{3}, Clément Mallet\textsuperscript{4}, Christian Heipke\textsuperscript{5} }
%\author{***** (for review, names must be rendered anonymous)}
 \author{Tomislav Medic\textsuperscript{1}, Liangliang Nan\textsuperscript{2}}
% KAO: Remove extra newline
% Anonymous submissions, authors' affiliations should not be visible
\address{
	\textsuperscript{1 }Institute of Geodesy and Photogrammetry, ETH Zurich, Zurich, Switzerland – tmedic@ethz.ch\\
	\textsuperscript{2 }Urban Data Science Section, Delft University of Technology, Delft, Netherlands – Liangliang.Nan@tudelft.nl\\
%	\textsuperscript{3 }Université Côte d’Azur, INRIA – Sophia-Antipolis, France – florent.lafarge@inria.fr\\
%	\textsuperscript{4 }Univ. Gustave Eiffel, IGN-ENSG, LaSTIG – Saint-Mandé, France – clement.mallet@ign.fr\\
%	\textsuperscript{5 }Institute of Photogrammetry and GeoInformation, Leibniz Universit\"at Hannover, Germany - heipke@ipi.uni-hannover.de\\
}
%\address{**** (for review, affiliations must be rendered anonymous)}

% If the corresponding author is NOT the final author, always add a % space before the subsequent comma, i.e.
% first author name\textsuperscript{a,}\thanks{Corresponding author} , % second author name \textsuperscript{b}, etc.
% thanks to Niclas Borlin 05-05-2016
% information on the corresponding author should not be used any longer and has been commented out
% C. Heipke, Jan 03,2024

% the use of the information of commissions and working groups should not be used any longer and has been commented out
% C. Heipke, Sept. 20,2022
%\commission{XX, }{YY} %This field is optional. If filled, XX and YY should be replaced by adequate numbers. See https://www2.isprs.org/commissions/
%\workinggroup{XX/YY} %This field is optional.
%\icwg{}   %This field is optional.

% KAO: Use times symbol
\abstract{

3D instance segmentation for laser scanning (LiDAR) point clouds remains a challenge in many remote sensing-related domains. Successful solutions typically rely on supervised deep learning and manual annotations, and consequently focus on objects that can be well delineated through visual inspection and manual labeling of point clouds. However, for tasks with more complex and cluttered scenes, such as in-field plant phenotyping in agriculture, such approaches are often infeasible. In this study, we tackle the task of in-field wheat head instance segmentation directly from terrestrial laser scanning (TLS) point clouds. To address the problem and circumvent the need for manual annotations, we propose a novel two-stage pipeline. To obtain the initial 3D instance proposals, the first stage uses 3D-to-2D multi-view projections, the Grounded SAM pipeline for zero-shot 2D object-centric segmentation, and multi-view label fusion. The second stage uses these initial proposals as noisy pseudo-labels to train a supervised 3D panoptic-style segmentation neural network. Our results demonstrate the feasibility of the proposed approach and show performance improvementsrelative to Wheat3DGS, a recent alternative solution for in-field wheat head instance segmentation without manual 3D annotations based on multi-view RGB images and 3D Gaussian Splatting, showcasing TLS as a competitive sensing alternative. Moreover, the results show that both stages of the proposed pipeline can deliver usable 3D instance segmentation without manual annotations, indicating promising, low-effort transferability to other comparable TLS-based point cloud segmentation tasks.

}


\maketitle

%\saythanks % added 28-02-2014 Markus Englich

\section{Introduction}\label{sec:intro}

Automated in-field 3D plant phenotyping remains a challenge in agriculture, especially when aiming for high-throughput field phenotyping (HTFP). LiDAR, and terrestrial laser scanners (TLS) in particular, have emerged as promising sensing technologies that offer a good trade-off between throughput and geometric data quality~\cite{jin2021lidar,medic2023challenges}, providing a viable option for 3D HTFP. However, a crucial yet unsolved problem in the data processing chain is effective 3D instance segmentation of plant organs in point clouds, a necessary prerequisite for analyzing detailed plant morphology. This challenge becomes increasingly difficult to solve for crops with intricate canopies, such as wheat.
% and when trading data quality for throughput. 

Recent efforts of 3D instance segmentation of plant organs have mostly focused on point clouds derived from multi-view photogrammetric reconstruction using dedicated camera setups (not LiDAR), individual plants (rather than canopies), stem vs. leaf segmentation (without finer organ-level detail), and indoor data acquisition (opposed to in-field)~\cite{song2025comprehensive,jin2025deepreview}. When it comes to in-field acquired TLS point clouds, most efforts have focused on stem-leaf segmentation for maize plants~\cite{jin2019separating,jin2025deepreview} or tiller counting for wheat plants~\cite{gu2023comparison}. The former was targeted due to a relatively simple scene geometry (large plants and plant organs, clear spacing between plants), while the latter benefits from problem simplification and approximation (it is easier to roughly identify and count than to isolate complete instances for detailed morphological analysis). These promising first results have not yet been extended to broader 3D plant phenotyping or to more complex crops such as wheat.

Despite the undeniable importance of wheat and the relevance of wheat head morphology to key traits such as yield~\cite{hund2019non}, 3D wheat head instance segmentation has not yet been attempted using LiDAR (including TLS) point clouds. A few related works reported success using structured light and laser triangulation scanners~\cite{liu2023extraction,wang2022unsupervised}, which provide point clouds of exceptionally high resolution and quality. However, these technologies require specialized infrastructure (e.g., field phenotyping platforms), which limits their usability, and their throughput is comparably lower than that of LiDAR platforms such as TLS~\cite{paulus2019measuring}. Hence, there is considerable value in achieving comparable results using a LiDAR-based platform like TLS, enabling more effective HTFP. 

As an alternative to scanning-based solutions for in-field 3D wheat head instance segmentation, ~\cite{zhang2025wheat3dgs} presented Wheat3DGS, an approach using a dedicated field phenotyping platform with a multi-camera rig and 3D Gaussian Splatting based 3D scene representation \cite{kerbl20233d}. The approach demonstrated competitive performance in both throughput and quality and arguably presents the state-of-the-art solution for this particular phenotyping task. In this work, we use Wheat3DGS as a baseline and aim to match its performance, using TLS point clouds instead of multi-view RGB images.

Existing scanning-based solutions for 3D plant phenotyping typically tackle instance segmentation of plant organs either using supervised 3D deep learning, unsupervised machine learning (clustering) with geometry-based and hand-crafted rules, or a combination of both. However, these strategies are not directly transferable to 3D wheat head instance segmentation in TLS point clouds, either due to a lack of annotated data or due to large differences in point cloud characteristics, particularly resolution and quality~\cite{paulus2019measuring}. Hence, to solve this task, we extended our search for an appropriate solution beyond the agricultural domain.

In general, 3D instance segmentation has advanced rapidly across domains with different constraints. In robotics, typically characterized by indoor and camera-rich environments, many pipelines reduce the 3D problem to 2D by segmenting instances in images and then associating them across views back into 3D using multi-view consistency, as done in~\cite{yan2024maskclustering,he2025pointseg}. These pipelines can operate with minimal or no manual annotations by leveraging strong pre-trained (foundation) image models, alleviating the labeling bottleneck. However, they rely on dense coverage and may degrade under clutter, occlusion, repetitive structures, or limited overlapping views, conditions commonly encountered in HTFP.
%, e.g., with reprojection and IoU agreement. 
%This approach is, in fact, also the basis of the mentioned  Wheat3DGS~\cite{zhang2025wheat3dgs}. 

Alternative pipelines rely on supervised deep learning algorithms operating directly on 3D data (mostly point clouds). They can be primarily categorized into 3D bounding box proposals (common in autonomous driving) and learned point grouping (clustering) approaches (common in remote sensing)~\cite{xiang2023review}. Successful applications in remote sensing include airborne and terrestrial mobile mapping of urban areas and infrastructure, and tree-level instance segmentation in forestry~\cite{xiang2023review,xiang2024automated}. However, these approaches remain fully supervised, requiring labor-intensive, per-point 3D annotations that are difficult and expensive to obtain. This limits their applicability to cases where manual labeling is feasible, which is not the case for HTFP of complex canopies.
%which can happen if the scene complexity impairs visual analysis and delineating 3D objects from their surroundings.
%which can to a large degree be related to a 3D scene complexity and clutter.
%Both instance segmentation approaches (3D-to-2D projections and supervised 3D deep learning) to some degree struggle with the tasks such as in-field plant phenotyping of complex canopies, however, in different stages of data processing (manual annotation vs. 3D instance identification).

In this study, we tackle the problem of in-field 3D instance segmentation of wheat heads using TLS point clouds. To address it, we combine and adapt two complementary 3D instance segmentation strategies into a two-stage pipeline. We adapt a 3D-to-2D projection approach for TLS point clouds (thereof derived 2D image representations) and employ the Grounded SAM framework~\cite{ren2024grounded} to eliminate the need for manual annotations. This forms Stage-I of our pipeline, which already provides promising results. To further relax the requirement for strict multi-view consistency, which is difficult to guarantee in cluttered or complex canopies, we introduce Stage-II, where Stage-I outputs are used as pseudo-labels to train a 3D panoptic-style deep neural network. Although these labels are noisy and contain missed detections, we hypothesize that the network can still learn the discriminative 3D features of wheat heads and recover them even in cluttered areas. Our results confirmed this hypothesis. 

To verify the practical relevance of our results for wheat HTFP, we compared our approach with Wheat3DGS. The comparison demonstrates that our TLS-based pipeline provides a viable and competitive alternative to the state of the art in in-field 3D wheat head instance segmentation and phenotyping.

% The results of the comparison cannot be interpreted "directly", as the method works on different data-modality (multi-view RGB) images. However, we chose the method as it is the only one that also allows for 3D instance segmentation of in-field observed wheat heads without manual 3D annotations, and is successfully applied on the same dataset as the one used in this study. 
% Irrespective of challenges arising from these differences, the results clearly demonstrate that the realized instance segmentation is of comparatively higher quality, marking it as a relevant contribution for this use-case.

%The article is organized as follows: Section 1 presented introduction with a brief review of relevant state of the art, problem statement and presentation of the proposed solution; Section 2 presents the implemented two-stage pipeline for 3D instance segmentation; Section 3 presents the experimental setup used to evaluate the implemented pipeline; Section 4 presents our first results and related discussion, while the main conclusions are drawn in Section 5.

\section{Implemented 3D Instance Segmentation Pipeline}\label{sec:method}

Our 3D instance segmentation pipeline comprises two stages: a multi-view projection-based stage (Stage-I, Sec.~\ref{sec:tls2dseg}) and a panoptic-style neural network stage (Stage-II, Sec.~\ref{sec:binbin}). The pipeline is general purpose and does not incorporate any task-specific priors for wheat head segmentation.

\subsection{Multi-View Projection-Based 3D Instance Segmentation}\label{sec:tls2dseg}

\begin{figure}[ht!]
\begin{center}
		\includegraphics[width=1.0\columnwidth]{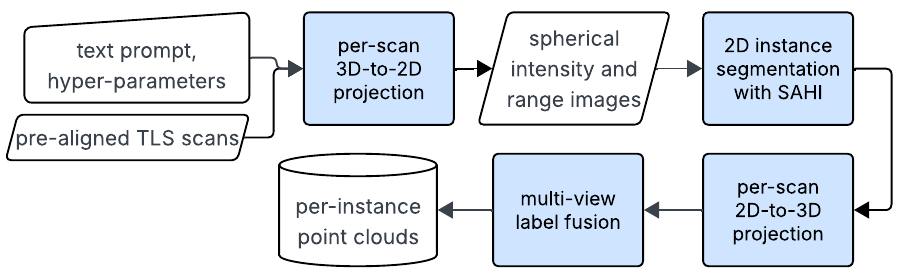}
	\caption{Flowchart of the implemented multi-view projection-based 3D instance segmentation (white: I/O, blue: data processing modules).}
\label{fig:flow_chart}
\end{center}
\end{figure}

The Stage-I of our pipeline uses a multi-view projection-based 3D instance segmentation strategy that follows the mentioned works on indoor 3D scene understanding in robotics, e.g., see~\cite{yan2024maskclustering,he2025pointseg}. The overview of Stage-I is provided in Fig.~\ref{fig:flow_chart}. The necessary input consists of: (i) per-scan-station (multi-view) TLS point clouds (XYZ+intensity values) that were previously aligned in a common local coordinate system (LCS), (ii) a text prompt defining the objects of interest within the scene, i.e., the targets of the 3D instance segmentation, and (iii) hyper-parameters that are steering the behavior of the pipeline. The main hyper-parameters are: $d_p$ - the desired point spacing (resolution) of the output point cloud and $r_{max}$ - the maximal range $r$ of interest (explanation follows).

% RoI - region of interest within the scene defined as a 2d polygon within the LCS,
% auto-detection of scan-resolution; auto-resolving orientation around z-axis; auto-flipping upside-down scans to reduce issues on poles;

\begin{figure}[ht!]
\begin{center}
		\includegraphics[width=1.0\columnwidth]{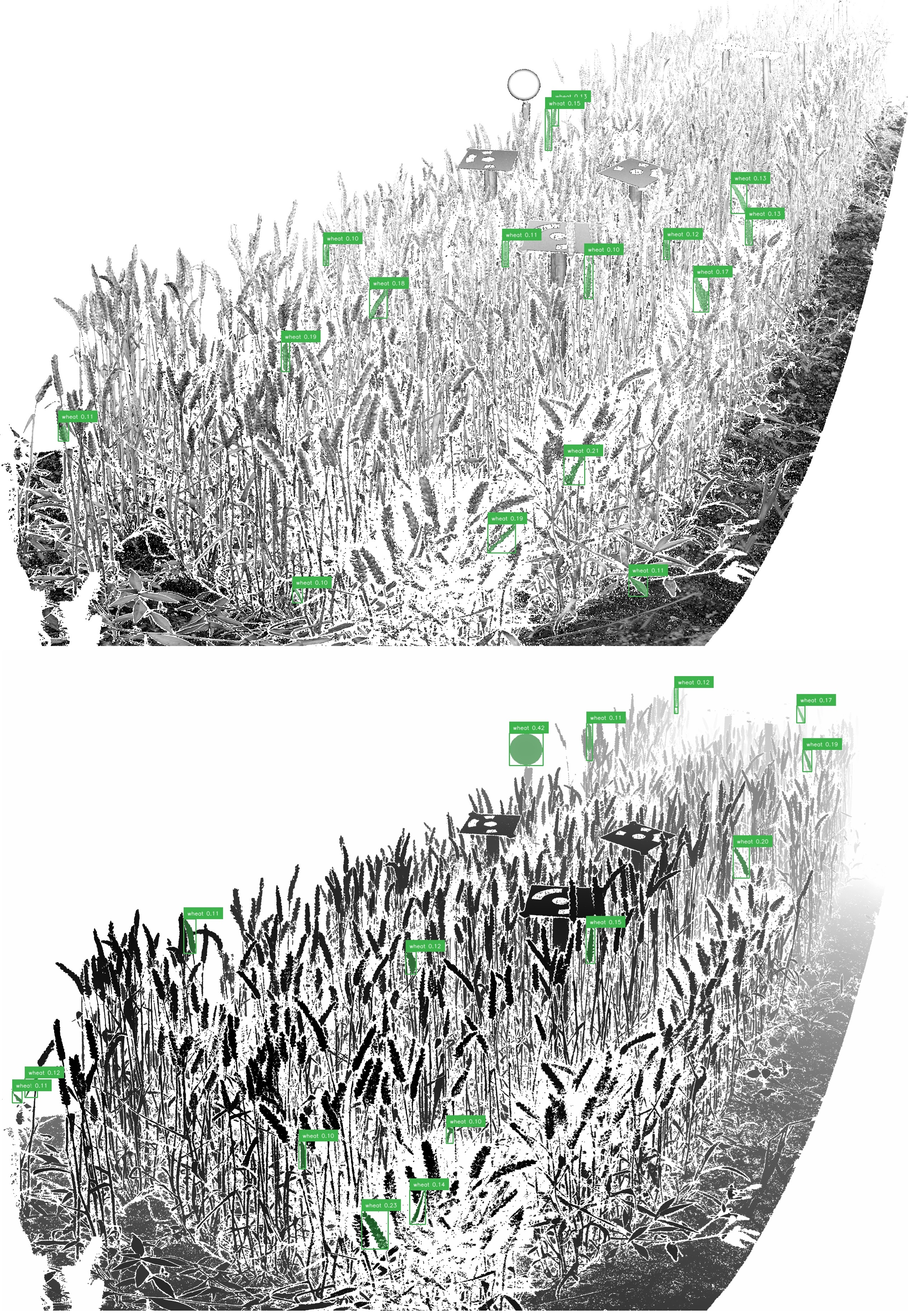}
	\caption{Examples of intensity (top) and range (bottom) images generated by the per-scan 3D-to-2D projection module, with a few examples of 2D instance masks generated by the 2D instance segmentation module. For clarity, only a small subset of the masks is shown.}
\label{fig:intensity_images}
\end{center}
\end{figure}

The first module (Fig.~\ref{fig:flow_chart}, blue boxes) projects the per-station point clouds into native spherical (panoramic) range and intensity monochromatic images (x- and y- image axes correspond to horizontal and vertical angle measurements). Initially this is done losslessly with a resolution corresponding to the scan resolution. Subsequently, the image resolution is reduced by image compression using Lanczos resampling algorithm~\cite{duchon1979lanczos} to a target ground sampling distance of $d_p$ at the maximum range $r_{max}$. This reduces the computational burden and facilitates efficient image processing in the following steps. An example of the resulting intensity and range images is shown in Fig.~\ref{fig:intensity_images}.
% RASTERIZATION: NAN-COV -> if pixel has value, it takes that value, otherwise Gaussian Kernel interpolation

These images are fed into the 2D instance segmentation module which relies on Grounded-SAM~\cite{ren2024grounded} inference pipeline, which incorporates two foundational image-processing models: Grounding DINO~\cite{liu2024grounding} - an open-set object detector steered by an arbitrary textual prompt as input, and Segment Anything Model (SAM)~\cite{kirillov2023segment} which can transform object detections (bounding boxes) into high-quality instance segmentation masks. We adopted the Grounded-SAM implementation which substitutes SAM, with SAM2~\cite{ravi2024sam} for increased mask quality and incorporates Slicing-Aided-Hyper-Inference or SAHI~\cite{akyon2022slicing} to allow for processing large panoramic images, which essentially breaks large images into smaller overlapping image patches, runs inference on individual patches, and combines the results using the Non Maximum Suppression (NMS). 

As the text prompt is used to define the objects of interest within the scene (prompt example: "house. window. car."), the output of this module is per-instance 2D masks paired with corresponding semantic classes for each of the input images (two images per scan - range and intensity, and one class per an object defined within the text prompt). We do not fuse the segmentation results of the corresponding intensity and range images, e.g., using NMS, but keep them as independent observations of the scene (i.e., independent views), as they refer to different observation modalities and bring somewhat independent evidence of the observed instances. The obtained 2D instance segmentation masks are projected back into the originally acquired 3D points in LCS, using known point-to-pixel correspondences. Each resulting 3D instance represents only a partial view of an object, with potentially corresponding detections across multiple views (scan-stations) or multiple modalities (intensity and range images).

The 3D instance correspondence search and multi-view + multi-modal label fusion (Fig.~\ref{fig:flow_chart}, last module) relies on a graph-based clustering technique that was largely inspired by MaskClustering~\cite{yan2024maskclustering} and adapted for our use case. We build a sparse undirected weighted connectivity-graph \(G=(V,E)\) where each node or vertex \(i,j,l,...\in V\) is a per-3D-instance oriented bounding box \(B\); edges E are first given by \(k\)-nearest neighbors considering $d(B_i, B_j)$ centroid distances in the Euclidean space and then pruned by the intersection over union (IoU) using a threshold $\tau$, so
\begin{equation}\label{equ:1}
(i,j)\in E \iff (i,j)\in E_{\text{kNN}}\ \land\ \operatorname{IoU}(B_i,B_j)\ge \tau. 
\end{equation}
For each retained edge \((i,j)\), the weight is the number of ``supporter'' nodes $l$ that have valid edges with both elements of the inspected edge \((i,j)\), i.e., it is the number of masks $B_l$ that have an $\operatorname{IoU} \ge \tau\ $ with both $B_i$ and $B_j$, which can be expressed as:
\begin{equation}\label{equ:2}
w_{ij}=\bigl|\{\,l\in V\setminus\{i,j\}:\ (i,l)\in E \ \land\ (j,l)\in E\,\}\bigr|.
\end{equation}
Hence, in our implementation $w_{ij}$ represents the count of triangles \((i,j,l)\) in the pruned graph, which is a simplification relative to the MaskClustering approach~\cite{yan2024maskclustering}. The clustering of the nodes is then realized using the Highly Connected Sub-graphs (HCS) algorithm~\cite{hartuv2000clustering}. The graph-based clustering output is a set of merged 3D instance point clouds obtained by uniting the instances associated with nodes \(\{i,j,l,\dots\}\in C\) for each cluster \(C\). The merged point clouds are then subsampled to the target point spacing \(d_p\) and expressed in the LCS. The Stage-I yields a standalone 3D instance segmentation result that serves both as input to Stage-II and as another baseline solution for comparison in Sec.~\ref{sec:results}.

Remark: Some of the implementation details are omitted for brevity and to adhere to the demands of this science communication format. They primarily comprise: a full list of hyperparameters, details on implemented outlier removal strategies, and details on auxiliary point cloud processing steps. In short, we implemented outlier removal strategies at several locations within the pipeline relying on either data-driven or a priori defined thresholds. The implemented point cloud processing steps mostly support generating spherical images and oriented bounding boxes with higher quality. Collectively, these steps help improve the computational efficiency and the quality of the results, but are non-essential for the functioning of the presented pipeline. Interested readers can find the details in the following public GitHub repository: https://github.com/tomedic/tls2dseg.

\subsection{Instance Segmentation using a Panoptic-Style Neural Network}\label{sec:binbin}

Extending our pipeline with Stage-II can be viewed as a task of 2D-to-3D label transfer and knowledge distillation, where we transfer the knowledge of 2D foundational teacher models (Grounding DINO and SAM2) to a model operating in 3D, but with a simultaneous distillation, focusing only on the objects of interest defined by the text-prompt provided as an input in Stage-I. The Stage-II adopts an algorithm for 3D panoptic segmentation for large-scale LiDAR point clouds introduced in~\cite{xiang2023towards}. We chose this algorithm, as it was proven to be effective for 3D instance segmentation in large-scale mobile mapping point clouds (similar to TLS), and effective for tree instance segmentation in forestry domain (similar to HTFP in agriculture). In this section, we provide a brief description of the algorithm, and the description of the interface between Stage-I and -II of our pipeline (i.e., producing the input for deep learning and generating noisy pseudo-labels for training).

\textbf{Algorithm.} The input point clouds are converted into regular sparse voxel grids and processed with a U-Net-style backbone built on Minkowski Engine~\cite{choy20194d} for feature extraction. On top, there are three MLP heads: (i) a semantic segmentation head trained with cross-entropy; (ii) a PointGroup-style instance clustering head~\cite{jiang2020pointgroup}; and (iii) a discriminative-embedding–based instance clustering head~\cite{de2017semantic}. Semantic predictions act as a logical gate to restrict instance search only to points related to preselected semantic classes. The two instance segmentation heads produce independent cluster proposals that are scored by a lightweight U-Net (``ScoreNet''), as in~\cite{jiang2020pointgroup}. Final instances are obtained by thresholding the predicted scores and applying the NMS. Because scenes (point clouds) are too large to process at once, they are sampled by overlapping spheres of a fixed radius and resolution for training and inference. These sphere-level outputs (model predictions) are merged into a unified result using threshold-based rules. For implementation details, we strictly follow~\cite{xiang2023towards} and its respective public codebase. Deviations are limited to dataset-specific hyper-parameters, out of which, the most relevant ones are listed in Sec.~\ref{sec:experiment}.

\textbf{Interface of Stage-I and -II.} We build the unlabeled target point cloud $P_t$ by merging all per-scan-station point clouds in the LCS and subsampling it to spacing $d_p$. We merge all 3D instance point clouds (Stage-I output) into a query point cloud $P_q$. Semantic and instance labels are transferred from $P_q$ to $P_t$ via nearest-neighbor interpolation with radius $d_p$, where points in $P_t$ without a neighbor from $P_q$ within a radius $d_p$ receive null labels for both semantic class and instance ID, and are marked as background. After transfer, labels are refined by: (i) weighted majority voting within a radius $d_p \cdot 1.2 \cdot \sqrt{3}$ ($\approx$ 3D voxel reachability with a small jitter margin of 1.2), and (ii) strict instance-wise majority voting: if $\ge 80\%$ of points in an instance share the same semantic class, all points in that instance get that class label; otherwise the instance is reset to null labels. This produces $P^t_q$, which serves as pseudo-labeled input for Stage-II.

At inference, the network predicts semantic and instance labels for all points in $P_t$; during training and evaluation (testing) it predicts on $P^t_q$, where the predicted labels are compared to the initial pseudo-labels for computing loss terms and evaluation metrics. We report the merged scene-level predictions (not per sampled sphere) and compare them against Stage-I outputs and our baseline (Wheat3DGS) in Sec.~\ref{sec:results}. Although we use an algorithm for panoptic segmentation, we do not claim panoptic segmentation with our implemented pipeline, as our pipeline provides exactly one non-informative ``stuff'' (uncountable, without instances) semantic class, the background class. Hence, the algorithm output reduces to 3D instance segmentation, and we report instance-level metrics accordingly (Sec.~\ref{sec:results}).

\section{Experiment}\label{sec:experiment}

In Sec.~\ref{sec:data}, we present the dataset used to test the implemented pipeline on the targeted use case of in-field 3D wheat-head instance segmentation. Additionally, we briefly explain how the baseline Wheat3DGS results are generated. In Sec.~\ref{sec:implementation}, we summarize the main data processing details, primarily main hyper-parameter choices and dataset-specific implementation details related to our pipeline.

\subsection{Data Collection}\label{sec:data}

\begin{figure*}[htbp!]
\begin{center}
		\includegraphics[width=0.9\textwidth]{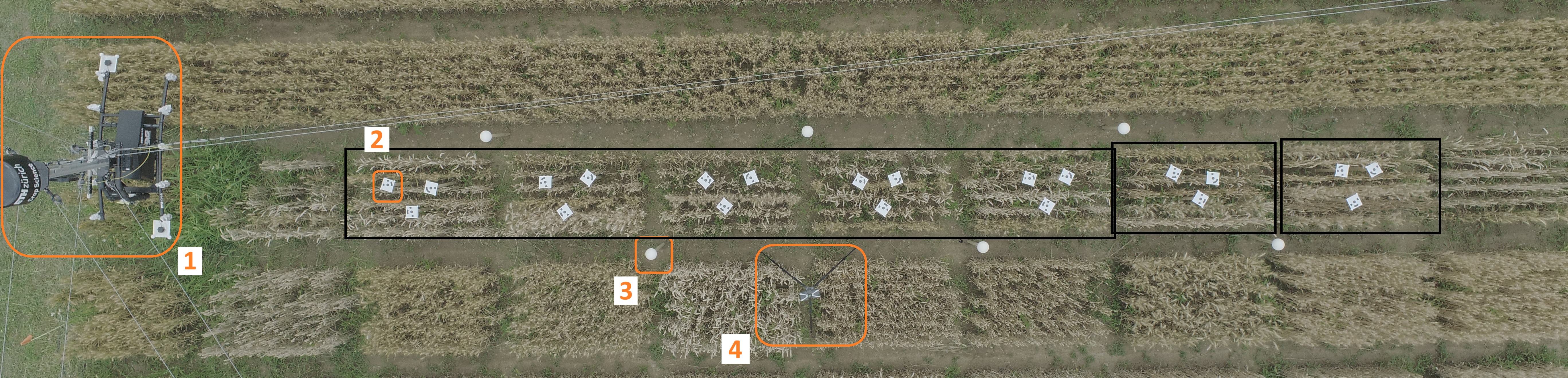}
	\caption{Overview of experiment setup and data collection. Orange rectangles: (1) field phenotyping platform with a camera rig, (2) photogrammetric target (coded marker), (3) reference sphere, (4) TLS. Black rectangles: 7 wheat plots covered within a measurement campaign, from left to right - 5 training, 1 validation, 1 test / hold-out plot.}
\label{fig:experiment_setup}
\end{center}
\end{figure*}

Our study uses an open-access dataset captured for the development and validation of the Wheat3DGS algorithm~\cite{zhang2025wheat3dgs}, a baseline solution used in Sec.~\ref{sec:results}. The dataset comprises multi-sensor observations (RGB images and TLS scans) of a small-scale wheat phenotyping experiment with 7 plots, each measuring approx. 1.5 m\textsuperscript{2} (Fig.~\ref{fig:experiment_setup}). The plots contain wheat plants with 42 different genetic varieties to induce variability and provide a minimal assessment of generalizability of the results. Each plot contains several hundred wheat heads, resulting in a few thousand instances overall.

The data used for Wheat3DGS consists of RGB images acquired by the Field Phenotyping Platform (FIP) \cite{Kirchgessner2017}, which carries a multi-view camera rig equipped with 13 cameras (12~MP). Image-based 3D reconstruction of each of the 7 plots was done using 36 images and was supported by 3 photogrammetric targets (coded markers) per plot to aid Structure from Motion (SfM), set the scale, and enable alignment with TLS scans. Based on this data, Wheat3DGS does 3D scene reconstruction using 3D Gaussian Splatting~\cite{kerbl20233d}, and subsequently performs 3D instance segmentation relying on a pipeline which resembles our Stage-I: it uses YOLOv5 2D object detector specifically trained to detect wheat heads \cite{david2021global}, SAM to retrieve masks per detected instance; and 2D-to-3D projections, IoU scores and greedy correspondence search to solve the multi-view label fusion problem (further details in~\cite{zhang2025wheat3dgs}).

The data used for testing our pipeline consists of TLS scans that were obtained using a FARO Focus 3D S 120 (FARO Technologies, Inc., FL, USA) operating with full resolution (1.6mm @ 10m) and the highest quality setting (``1x'' = no internal point averaging). The scanning setup comprised multiple scan positions (19), including upside-down scans from a custom mount installed approximately 1~m above the canopy, and side-scans from a tripod at several meters distance to ensure a comprehensive coverage. Reference spheres (6~x 15~cm in diameter) were placed within the scene to facilitate scan registration. The scans were registered within a common local coordinate system (LCS) in the FARO SCENE 2022.1.0 software using a target- or sphere-based algorithm, with a mean alignment error (sphere centroids mismatch post registration) of 3~mm (max. 10~mm). Such per-scan-station point clouds expressed in the LCS served as an input to our pipeline.

Additionally, to allow for comparison between TLS scans and Wheat3DGS, the authors in ~\cite{zhang2025wheat3dgs} additionally aligned the registered scans with the image-based 3D scene reconstruction. The alignment of the corresponding wheat plants between the TLS and Wheat3DGS was assessed to be on average within 10~mm based on visual inspection, while we have observed the maximal misalignment of up to 30~mm. Further improvement in the alignment would require non-rigid transformations of the point clouds to compensate for plant motion~\cite{medic2023challenges}. However, this was not done due to a lack of readily available algorithms.

%Additionally, to allow for comparison between our results and Wheat3DGS, the registered scans were aligned with the image-based 3D scene reconstruction in two steps: first coarsely using the Kabsch algorithm~\cite{lawrence2019purely} and a sparse set of corresponding points identified in both datasets (using photogrammetric targets), then precisely using a standard implementation of the Iterative Closest Point (ICP) algorithm on subsampled point clouds (0.5 mil. points). The alignment of the corresponding wheat plants between the TLS and Wheat3DGS was assessed to be on average within 10~mm based on visual inspection (max. misalignment of up to 30~mm). Further improvement in the alignment would require non-rigid transformations of the point clouds to compensate for plant motion~\cite{medic2023challenges}. However, this was not done due to a lack of readily available algorithms.

\subsection{Data Processing}\label{sec:implementation}

As mentioned in Sec.~\ref{sec:tls2dseg}, the Stage-I requires 3 inputs: (i) TLS point clouds in a common LCS; (ii) a text prompt defining the instance segmentation targets, in this case a single word prompt ``wheat''; (iii) a set of hyper-parameters steering the Stage-I, where the main ones are $d_p$ of 3~mm, $r_{max}$ of 6~m, $\tau$ of 0.15, $k$ for knn equal to the number of scans (19).

Once the results of Stage-I are transferred to the merged and subsampled point cloud $P_t$, in Stage-II, this dataset requires solving a 2-class semantic segmentation (wheat vs. background) and an instance segmentation for the wheat class. To limit the overfitting and demonstrate some generalizability of the Stage-II, we spatially split the dataset according to the plots as shown in Fig.~\ref{fig:experiment_setup}, with 5 plots for training, 1 for validation and hyper-parameter tuning, and 1 for testing. For training and inference, we sample spherical point cloud samples (Sec.~\ref{sec:binbin}) of 0.12~m radius and $d_p$ resolution. During training, each sampled sphere passed through data augmentation (regularization strategy) that introduces randomized point cloud transformations consisting of: applying random noise (0 mean, 0.3~mm stdev), random rotations of up to 180 degrees about the Z-axis, random anisotropic scaling with a scale factor in the range of 0.9-1.1, and random flips about the X-axis. To facilitate efficient training, we used batches of 16 spheres and accordingly adjusted the learning rate to 0.004. All other parameters and settings (incl. e.g., the choice of the optimizer, number of epochs, and learning rate schedule) were kept as in the original publication and can be found in the related codebase. Training was run on a workstation with an NVIDIA GeForce RTX 3090 Ti (24 GB RAM, CUDA 11.5) and took ~12 hours. Remarks: Parameters $d_p$, $r_{max}$, and sphere radius are set empirically, while the parameters $\tau$ and $k$ are chosen experimentally based on trial and error or educated guesses.
% based on our domain knowledge

\section{Results}\label{sec:results}

\begin{figure}[ht!]
\begin{center}
		\includegraphics[width=1.0\columnwidth]{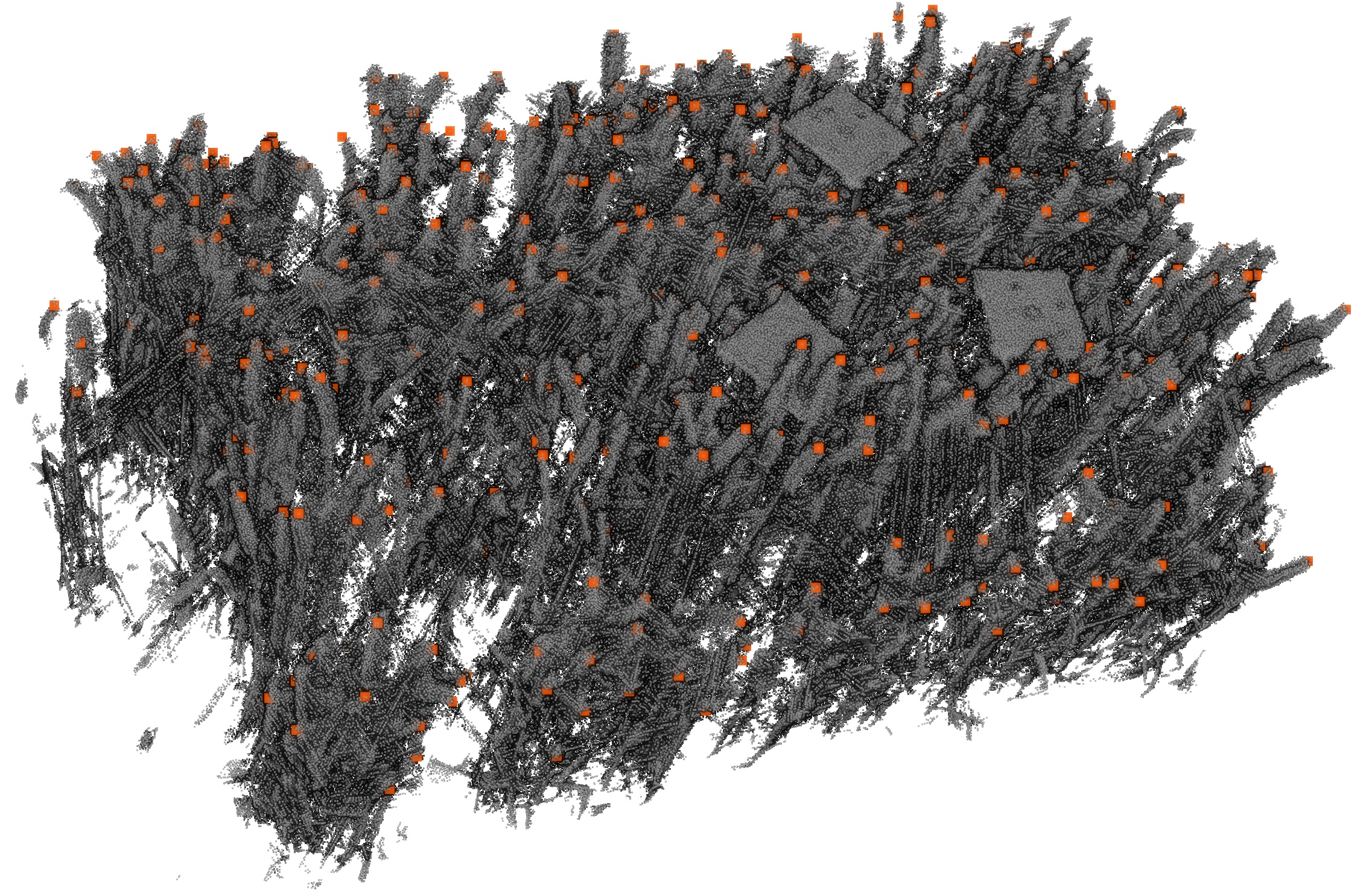}
	\caption{Sparse reference 3D instance annotations (red) overlaid over $P_t$ point cloud of the test / hold-out plot.}
\label{fig:gt}
\end{center}
\end{figure}

We evaluated the quality of the 3D wheat head instance segmentation against a set of sparse reference labels. Due to the high occlusion and complexity of the canopy, producing a dense per-point ground truth was infeasible. Instead, for each visually identifiable wheat head, we manually annotated a single 3D point \(g_j\), yielding a sparse set \(G\) where \(|G|\) equals the total number of wheat heads in the scene. Thus, each \(g_j\in G\) should lie within one predicted wheat-head instance in \(P_t\). All metrics are computed only for the test / hold-out plot (Fig.~\ref{fig:experiment_setup}), which was not used for training in Stage-II. The related point cloud \(P_t\) along with the corresponding reference annotations \(g_j\) are presented in Fig.~\ref{fig:gt}.

For quantification, we computed the following values. For each segmented 3D instance \(I_k\subset P_t\) (with points \(p\in I_k\)) and each reference point \(g_j\in G\), we compute a spatial matching distance
\(d_{jk}=\min_{p\in I_k}\lVert p-g_j\rVert_2\).
With a fixed distance threshold \(d_\tau=30\,\mathrm{mm}\), we solve a one-to-one assignment using the Hungarian algorithm~\cite{kuhn1955hungarian} on the matrix $D_{jk}$, disallowing matches with \(d_{jk}>d_\tau\). Matched pairs are considered true positives (TP), unmatched predictions as false positives (FP), and unmatched references as false negatives (FN). We report $F1$ score, precision, and recall next to the TP/FP/FN counts. For completeness, we also report absolute and relative counting errors (CE and RCE) relative to the number of manual reference detections.

\begin{figure}[ht!]
\begin{center}
		\includegraphics[width=1.0\columnwidth]{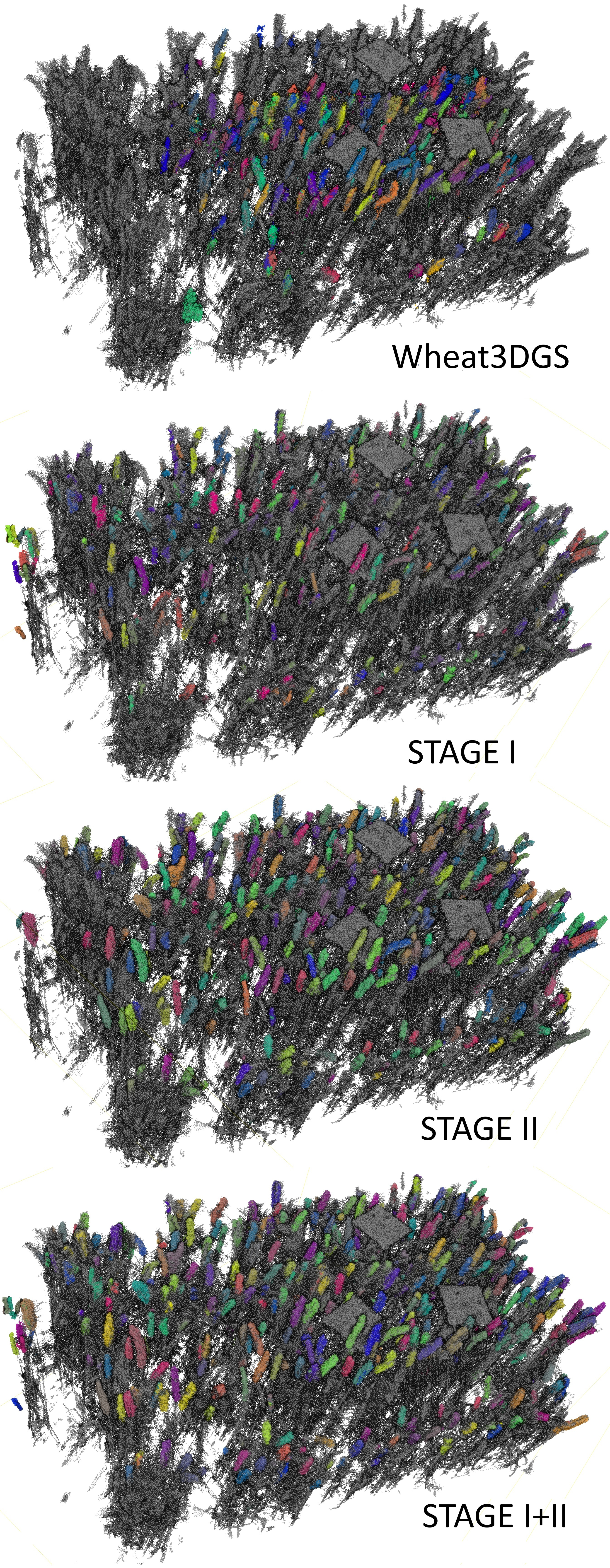}
	\caption{Detected 3D wheat head instances (random color per instance alignment) overlaid over $P_t$ point cloud of the hold-out (test) plot. A few side views are shown in Appendix A.}
\label{fig:results_visual}
\end{center}
\end{figure}

We computed those metrics for four 3D wheat head instance segmentation realizations: (1) Wheat3DGS as a baseline approach, (2) Stage-I results as own possible independent solution, (3) Stage-II results as the intended end result of our pipeline, but also (4) the joint results of Stage-I and Stage-II (merged using NMS with 3D IoU, with 0.1 threshold),  All results are summarized in Tab.~\ref{tab:results} and the corresponding data is illustrated in Fig.~\ref{fig:results_visual}. The analysis of the results is separated in the following three subsections.
% threshold chosen based on trial and error
\begin{table}[h]
	\centering
	\begin{tabular}{|l|c|c|c|c|}\hline
		Metric & W3DGS & Stage-I & Stage-II & I+II \\\hline
		TP  & 83  & 267 & 296 & 348 \\
		FP  & 228 & 109 & 103 & 165 \\
		FN  & 375 & 191 & 162 & 110 \\
		P   & 0.27 & 0.71 & 0.74 & 0.68 \\
		R   & 0.18 & 0.58 & 0.64 & 0.76 \\
		F1  & 0.22 & 0.64 & 0.69 & 0.72 \\
		CE  & -147 & -82 & -59 & 55 \\
		RCE & -0.32 & -0.18 & -0.13 & 0.12 \\\hline
	\end{tabular}
	\caption{3D instance segmentation evaluation metrics for Wheat3DGS (W3DGS) and our pipeline; separately for Stage-I, -II and combined I+II. Legend: TP - true positives, FP - false positives, FN - false negatives, P - precision, R - recall, F1 - F1 score, CE - counting error, RCE - relative counting error, USR - under segmentation ratio in percent.}
\label{tab:results}
\end{table}

% Remarks: Reference annotations may include missed detections and occasional false positives due to observer errors. Moreover, Wheat3DGS predictions rely on a different base dataset (Sec. 3.1) and they required alignment to TLS. Hence, residual alignment error may negatively affect its scores. We mitigate this by using a generous correspondence threshold \(d_\tau\) of 30~mm (chosen based on the worst observed wheat head alignment error, see Sec. 3.2). Therefore, the reported values should be interpreted as indicative rather than definitive. They, however, agree with visual inspection, and we deem them as sufficient to support our conclusions.
Remark: We mitigate potential misalignment issues by using a generous correspondence threshold \(d_\tau = 30~mm\), which was chosen to exceed the maximum observed alignment error between the TLS and Wheat3DGS datasets (Sec.~\ref{sec:data}). While the sparse nature and limited quality of the manual reference annotations should be considered, the reported metrics are consistent with visual inspection and provide a sufficiently robust quantitative basis for comparing the methods.

\subsection{Comparison Against the Benchmark}\label{sec:results_benchmark}

%The results in Tab.~\ref{tab:results} demonstrate that our TLS-based pipeline (any solution) surpasses the performance of the image-based Wheat3DGS pipeline, with the F1-score more than tripling.
The data in Tab.~\ref{tab:results} suggests that our TLS-based pipeline (any solution) outperforms the image-based Wheat3DGS baseline on the selected metrics, best indicated by the noticeably higher F1-score. This showcases TLS, in conjunction with our method, as a competitive sensing modality for 3D high-throughput field phenotyping. Pinpointing exact causes for the divergence between Wheat3DGS- and TLS-based results is beyond the scope of this work. It is possible that it is partially caused by a superior distribution of TLS viewpoints relative to the distribution of camera viewpoints used for Wheat3DGS. Consequently, it is unclear which method would overperform under unbiased conditions, and, hence, the reported differences in the performance should in this case be interpreted as indicative rather than definitive. Nevertheless, what we want to highlight is that our method is evidently effective. It at least matches and potentially further elevates the bar relative to a very recent SOTA technical solution for in-field wheat head 3D instance segmentation and phenotyping. 

On one hand, our solution presents a viable, and rather easily re-usable methodology for a specific agriculture domain problem, namely the in-field assessment of wheat head morphology at scale (with high throughput), where 3D instance segmentation is a necessary prerequisite. The presented solution does require a rather costly instrument (TLS), but does not require a dedicated permanently installed infrastructure, like common field phenotyping platforms, as used e.g., in ~\cite{liu2023extraction}. Moreover, the existence of such a TLS-based technical solution offers, if not a reference, at least an independent control that can be used for further development and evaluation of the image-based approaches for 3D phenotyping and morphology analysis at scale. In many other domains, this has proven itself to be a cornerstone of efficient algorithmic development, where TLS point clouds are considered to be an undisputed reference for 3D scene geometry.

On the other hand, our results demonstrate that it is possible to solve rather challenging cases of 3D instance segmentation in TLS (LiDAR) point clouds mostly by relying on the existing 2D image domain foundational models, circumventing the need for manual annotations. In other words, our results demonstrate a zero-shot solution for 3D data. This is especially important for similar use cases where creating dense ground truth is infeasible or prohibitively costly. In our case, for example, it is virtually infeasible for a human annotator to solve this annotation problem, mostly due to a high level of occlusions in a complex canopy structure and a high ratio of erroneous and noisy points that stem from mixed pixels \cite{medic2023challenges}. For any comparable task, our pipeline might present a viable way forward. Moreover, as the method is a zero-shot method, our results promise a potential for a rather broad applicability and generalizability of the implemented pipeline. However, confirming that will be a part of future efforts. 

\subsection{Comparison of Stage-I and -II}\label{sec:results_stages}

% On one hand, that means that the Stage-I of our pipeline, the multi-view projection-based 3D instance segmentation method, offers a strong zero-shot baseline solution. On the other hand, it shows that the intended cross-domain (2D to 3D) knowledge transfer and distillation was successful. Using our pipeline and the Stage-I predictions as noisy pseudo-labels, it was possible to train a 3D neural-network based classifier relying only on the knowledge of 2D foundational models.
% Solely by comparing the metrics of Stage-I and -II, it is hard to assess if there is any clear benefit of introducing Stage-II in our pipeline at all. However, through visual inspection we have noticed that 

While the performance of Stage-I and Stage-II is comparable, Stage-II provides a measurable improvement in all evaluation metrics (Tab.~\ref{tab:results}). More importantly, visual inspection revealed that some of the wheat heads that were not detected in Stage-I, were detected in Stage-II and vice versa (Fig.~\ref{fig:results_visual} and Fig.~\ref{fig:a1}). This is not surprising, as our Stage-I has a hard requirement of having at least 3 independent observations of a single instance for it to be accepted as a valid 3D instance (see Sec.~\ref{sec:tls2dseg}). This is a requirement that can be hard to satisfy for all wheat heads due to the heavy occlusions. However, the segmentation model in Stage-II learns from Stage-I examples to recognize wheat heads in 3D, independent of the number of observations, and is capable of detecting instances that were initially missed (e.g. regions 2 and 3 in Fig.~\ref{fig:a1}). Unfortunately, it also misses some instances that were detected by Stage-I (e.g. regions 4 and 5 in Fig.~\ref{fig:a1}). This can be quantified and observed if we compare the true positives and recall values for all solutions, including the combined results of Stages I- and II (I+II). If instances detected by both stages are combined, we can find noticeably more wheat heads, which is best reflected in the true positives count, recall and F1-score in Tab.~\ref{tab:results} (also visible in Fig.~\ref{fig:results_visual} and Fig.~\ref{fig:a1}). Hence, we conclude the following: (i) introducing Stage-II does have its merits; (ii) both Stage-I and -II have their own advantages; (iii) combining the results of both might provide the best 3D instance segmentation (and counting) performance.
% \LL{I have a concern: each of the stages may produce errors (false positives), and combining the results of them also leads to a combined error. It will be more convincing to observe that Stage 2 also **corrects** some errors from Stage 1, thereby reducing the overall error. Can you do some analysis on this?}

\subsection{Analysis of Failure Cases}\label{sec:results_failures}

By visual inspection, we identified some common instance segmentation problems. For once, as anticipated, Stage-II has difficulty coping with out-of-the-distribution samples, e.g., with rare wheat heads that are very large, curved, and hanging more horizontally than vertically. This partially explains why Stage-II is not clearly superior to Stage-I regarding metrics in Tab.~\ref{tab:results} (Stage-I is more immune to this problem). Addressing it either requires a deployment of a suitable strategy, e.g., identification of such cases and corresponding data augmentation, or it can be partially resolved by our ``hotfix'', by combining Stage-I and -II predictions.

Also, most of the missed detections (FN) can be related to very cluttered portions of the scene, having multiple ($\geq 5$) wheat heads forming a dense cluster. In these cases, correctly discerning individual wheat heads can be overly challenging for the proposed pipeline (again primarily for Stage-II), which sometimes results in under-segmentation, i.e., having multiple wheat heads being assigned to one 3D instance (e.g. region 3 in Fig.~\ref{fig:a1}). To confirm that we computed an under segmentation ratio $USR$ as follows: for each segmented instance \(I_k\), we search for reference labels \(G_k=\{\,g_j\in G:\ d_{jk}\le d_\tau\,\}\). If the condition \(|G_k|\ge 2\) is fulfilled, \(I_k\) is considered to be under-segmented, i.e., it likely contains more than one wheat head within one 3D instance $I_k$. For this analysis, we focus on our pipeline and skip the analysis of Wheat3DGS results. As a result, we deploy a stricter $d_\tau$ of only 10~mm (instead of the initial 30~mm). The corresponding $USR$ for Stage-I -II and -I+II were 2.4\%, 9.5\%, and 7.5\% respectively. These values confirm our claims for Stage-II results, as this $\approx$ 10\% $USR$ value accounts for a substantial portion of the false negatives ($\approx$ 30\%) and the counting error (CE and RCE in Tab.~\ref{tab:results}). This issue may be mitigated by improved post-processing, particularly in the NMS step and the strategy for merging overlapping sphere predictions (see Sec.~\ref{sec:implementation}). An in-depth analysis of this bottleneck is left for future work. In the case of Stage-I, the large portion of false negatives and the counting errors are rather linked to the aforementioned strict multi-observation requirement, which cannot be satisfied for all instances. Again, our claims are supported by the fact that the combined I+II solution provides better metrics than the individual Stage-I and -II solutions, indicating different causes for the detection errors.

Furthermore, a part of the drop in the Tab.~\ref{tab:results} values relates to the quality of our ground truth, especially the false positive values can be impacted, if our pipeline managed to identify the instances that the human annotators missed. Finally, some errors, especially those related to the combined I+II results, relate to overly simplified data-processing strategies. Namely, I+II results have too many overall detections (CE and RCE), false positives, and they exhibit a corresponding drop in precision, i.e., they indicate detecting many more wheat heads than are present within the scene. This can, to a large degree, be attributed to a naive Stage I and II results merging strategy, which sometimes fails to merge corresponding instances, introducing two detections per one ground-truth label, which in turn introduces false positives. However, further improvements are left for future work. 

% We would like to stress out that these zero shot pipeline outputs are of competitive quality. However, we cannot answer to which degree would they keep up with a dedicated solution relying on annotated ground truth, as such a solution did not exist for the use case that we tackled in this study.

\section{Conclusion}\label{sec:conclusion}

This study presented a novel, two-stage pipeline for zero-shot 3D instance segmentation in TLS point clouds, effectively eliminating the dependency on manual annotation. Our approach synergistically combines (i) a multi-view projection-based stage that leverages the zero-shot capability of foundational 2D models (Grounded-SAM) for initial instance proposal generation, with (ii) a second stage that distills this 2D knowledge into a supervised 3D panoptic-style neural network trained on pseudo-labels. We validated our method on the challenging agricultural task of in-field 3D wheat head instance segmentation. Our analysis on a hold-out test plot suggests that, under given experiment conditions, our TLS-based approach outperforms Wheat3DGS, a state-of-the-art image-based baseline, noticeably improving the F1-score. This compelling result underscores two key contributions: (i) it establishes TLS coupled with our processing pipeline as a highly competitive technology for high-throughput field phenotyping; and (ii) it demonstrates the feasibility of high-quality zero-shot 3D instance segmentation by effectively transferring knowledge from powerful 2D vision foundation modals to the 3D domain. 

We acknowledge two primary limitations of this study: the use of sparse reference labels for evaluation and the performance ceiling of the chosen baseline. Furthermore, the pipeline has identifiable bottlenecks, including residual under-segmentation in dense instance clusters, sensitivity to out-of-the-distribution samples, and dependence on the heuristics used for merging overlapping predictions. Our future work will focus on addressing these limitations by refining the post-processing and fusion strategies, and exploring human-in-the-loop or active few-shot learning to efficiently correct errors in challenging cases. We will also validate the generalizability of the pipeline across different plant phenotyping and remote sensing tasks. In summary, this work provides a significant step forward in both the domain of annotation-efficient 3D instance segmentation and the practical application of 3D remote sensing for high-throughput agricultural phenotyping.

{
	\begin{spacing}{1.17}
		\normalsize
		\bibliography{references} % Include your own bibliography (*.bib), style is given in isprs.cls

\begin{thebibliography}{xx}

\bibitem[Akyon et al., 2022]{akyon2022slicing}
Akyon, F.~C., Altinuc, S.~O., Temizel, A., 2022.
 Slicing aided hyper inference and fine-tuning for small object detection.
 \emph{2022 IEEE international conference on image processing (ICIP)}, IEEE, 966--970.

\bibitem[Choy et al., 2019]{choy20194d}
Choy, C., Gwak, J., Savarese, S., 2019.
 4d spatio-temporal convnets: Minkowski convolutional neural networks.
 \emph{Proceedings of the IEEE/CVF conference on computer vision and pattern recognition}, 3075--3084.

\bibitem[David et al., 2021]{david2021global}
David, E., Serouart, M., Smith, D., Madec, S., Velumani, K., Liu, S., Wang, X., Pinto, F., Shafiee, S., Tahir, I.~S. et~al., 2021.
 Global wheat head detection 2021: An improved dataset for benchmarking wheat head detection methods.
 {\em Plant Phenomics}.

\bibitem[De~Brabandere et al., 2017]{de2017semantic}
De~Brabandere, B., Neven, D., Van~Gool, L., 2017.
 Semantic instance segmentation with a discriminative loss function.
 {\em arXiv preprint arXiv:1708.02551}.

\bibitem[Duchon, 1979]{duchon1979lanczos}
Duchon, C.~E., 1979.
 Lanczos filtering in one and two dimensions.
 {\em Journal of Applied Meteorology (1962-1982)}, 1016--1022.

\bibitem[Gu et al., 2023]{gu2023comparison}
Gu, Y., Ai, H., Guo, T., Liu, P., Wang, Y., Zheng, H., Cheng, T., Zhu, Y., Cao, W., Yao, X., 2023.
 Comparison of two novel methods for counting wheat ears in the field with terrestrial LiDAR.
 {\em Plant Methods}, 19(1), 134.

\bibitem[Hartuv and Shamir, 2000]{hartuv2000clustering}
Hartuv, E., Shamir, R., 2000.
 A clustering algorithm based on graph connectivity.
 {\em Information processing letters}, 76(4-6), 175--181.

\bibitem[He et al., 2025]{he2025pointseg}
He, Q., Peng, J., Jiang, Z., Hu, X., Zhang, J., 2025.
 Pointseg: A training-free paradigm for 3d scene segmentation via foundation models.
 \emph{Proceedings of the IEEE/CVF International Conference on Computer Vision}, 2657--2667.

\bibitem[Hund et al., 2019]{hund2019non}
Hund, A., Kronenberg, L., Anderegg, J., Yu, K., Walter, A., 2019.
 Non-invasive field phenotyping of cereal development.
 \emph{Advances in breeding techniques for cereal crops}, Burleigh Dodds Science Publishing, 249--292.

\bibitem[Jiang et al., 2020]{jiang2020pointgroup}
Jiang, L., Zhao, H., Shi, S., Liu, S., Fu, C.-W., Jia, J., 2020.
 Pointgroup: Dual-set point grouping for 3d instance segmentation.
 \emph{Proceedings of the IEEE/CVF conference on computer vision and Pattern recognition}, 4867--4876.

\bibitem[Jin et al., 2025]{jin2025deepreview}
Jin, S., Li, D., Yun, T., Tang, J., Wang, K., Li, S., Yang, H., Yang, S., Xu, S., Cao, L. et~al., 2025.
 Deep learning for three-dimensional (3D) plant phenomics.
 {\em Plant Phenomics}, 100107.

\bibitem[Jin et al., 2019]{jin2019separating}
Jin, S., Su, Y., Gao, S., Wu, F., Ma, Q., Xu, K., Hu, T., Liu, J., Pang, S., Guan, H. et~al., 2019.
 Separating the structural components of maize for field phenotyping using terrestrial LiDAR data and deep convolutional neural networks.
 {\em IEEE Transactions on Geoscience and Remote Sensing}, 58(4), 2644--2658.

\bibitem[Jin et al., 2021]{jin2021lidar}
Jin, S., Sun, X., Wu, F., Su, Y., Li, Y., Song, S., Xu, K., Ma, Q., Baret, F., Jiang, D. et~al., 2021.
 Lidar sheds new light on plant phenomics for plant breeding and management: Recent advances and future prospects.
 {\em ISPRS Journal of Photogrammetry and Remote Sensing}, 171, 202--223.

\bibitem[Kerbl et al., 2023]{kerbl20233d}
Kerbl, B., Kopanas, G., Leimk{\"u}hler, T., Drettakis, G., 2023.
 3D Gaussian splatting for real-time radiance field rendering.
 {\em ACM Trans. Graph.}, 42(4), 139--1.

\bibitem[Kirchgessner et al., 2017]{Kirchgessner2017}
Kirchgessner, N., Liebisch, F., Yu, K., Pfeifer, J., Friedli, M., Hund, A., Walter, A., 2017.
 The {{ETH}} Field Phenotyping Platform {{FIP}}: {{A}} Cable-Suspended Multi-Sensor System.
 {\em Functional Plant Biology}, 44, 154--168.

\bibitem[Kirillov et al., 2023]{kirillov2023segment}
Kirillov, A., Mintun, E., Ravi, N., Mao, H., Rolland, C., Gustafson, L., Xiao, T., Whitehead, S., Berg, A.~C., Lo, W.-Y. et~al., 2023.
 Segment anything.
 \emph{Proceedings of the IEEE/CVF international conference on computer vision}, 4015--4026.

\bibitem[Kuhn, 1955]{kuhn1955hungarian}
Kuhn, H.~W., 1955.
 The Hungarian method for the assignment problem.
 {\em Naval research logistics quarterly}, 2(1-2), 83--97.

\bibitem[Liu et al., 2024]{liu2024grounding}
Liu, S., Zeng, Z., Ren, T., Li, F., Zhang, H., Yang, J., Jiang, Q., Li, C., Yang, J., Su, H. et~al., 2024.
 Grounding dino: Marrying dino with grounded pre-training for open-set object detection.
 \emph{European conference on computer vision}, Springer, 38--55.

\bibitem[Liu et al., 2023]{liu2023extraction}
Liu, Z., Jin, S., Liu, X., Yang, Q., Li, Q., Zang, J., Li, Z., Hu, T., Guo, Z., Wu, J. et~al., 2023.
 Extraction of wheat spike phenotypes from field-collected lidar data and exploration of their relationships with wheat yield.
 {\em IEEE transactions on geoscience and remote sensing}, 61, 1--13.

\bibitem[Medic et al., 2023]{medic2023challenges}
Medic, T., B{\"o}mer, J., Paulus, S., 2023.
 Challenges and recommendations for 3D plant phenotyping in agriculture using terrestrial lasers scanners.
 {\em ISPRS Annals of the Photogrammetry, Remote Sensing and Spatial Information Sciences}, 10, 1007--1014.

\bibitem[Paulus, 2019]{paulus2019measuring}
Paulus, S., 2019.
 Measuring crops in 3D: using geometry for plant phenotyping.
 {\em Plant methods}, 15(1), 103.

\bibitem[Ravi et al., 2024]{ravi2024sam}
Ravi, N., Gabeur, V., Hu, Y.-T., Hu, R., Ryali, C., Ma, T., Khedr, H., R{\"a}dle, R., Rolland, C., Gustafson, L. et~al., 2024.
 Sam 2: Segment anything in images and videos.
 {\em arXiv preprint arXiv:2408.00714}.

\bibitem[Ren et al., 2024]{ren2024grounded}
Ren, T., Liu, S., Zeng, A., Lin, J., Li, K., Cao, H., Chen, J., Huang, X., Chen, Y., Yan, F., Zeng, Z., Zhang, H., Li, F., Yang, J., Li, H., Jiang, Q., Zhang, L., 2024.
 Grounded sam: Assembling open-world models for diverse visual tasks.

\bibitem[Song et al., 2025]{song2025comprehensive}
Song, H., Wen, W., Wu, S., Guo, X., 2025.
 Comprehensive review on 3D point cloud segmentation in plants.
 {\em Artificial Intelligence in Agriculture}.

\bibitem[Wang et al., 2022]{wang2022unsupervised}
Wang, F., Li, F., Mohan, V., Dudley, R., Gu, D., Bryant, R., 2022.
 An unsupervised automatic measurement of wheat spike dimensions in dense 3D point clouds for field application.
 {\em Biosystems Engineering}, 223, 103--114.

\bibitem[Xiang et al., 2023a]{xiang2023towards}
Xiang, B., Peters, T., Kontogianni, T., Vetterli, F., Puliti, S., Astrup, R., Schindler, K., 2023a.
 Towards accurate instance segmentation in large-scale lidar point clouds.
 {\em arXiv preprint arXiv:2307.02877}.

\bibitem[Xiang et al., 2024]{xiang2024automated}
Xiang, B., Wielgosz, M., Kontogianni, T., Peters, T., Puliti, S., Astrup, R., Schindler, K., 2024.
 Automated forest inventory: Analysis of high-density airborne LiDAR point clouds with 3D deep learning.
 {\em Remote Sensing of Environment}, 305, 114078.

\bibitem[Xiang et al., 2023b]{xiang2023review}
Xiang, B., Yue, Y., Peters, T., Schindler, K., 2023b.
 A review of panoptic segmentation for mobile mapping point clouds.
 {\em ISPRS Journal of Photogrammetry and Remote Sensing}, 203, 373--391.

\bibitem[Yan et al., 2024]{yan2024maskclustering}
Yan, M., Zhang, J., Zhu, Y., Wang, H., 2024.
 Maskclustering: View consensus based mask graph clustering for open-vocabulary 3d instance segmentation.
 \emph{Proceedings of the IEEE/CVF Conference on Computer Vision and Pattern Recognition}, 28274--28284.

\bibitem[Zhang et al., 2025]{zhang2025wheat3dgs}
Zhang, D., Gajardo, J., Medic, T., Katircioglu, I., Boss, M., Kirchgessner, N., Walter, A., Roth, L., 2025.
 Wheat3dgs: In-field 3d reconstruction, instance segmentation and phenotyping of wheat heads with gaussian splatting.
 \emph{Proceedings of the Computer Vision and Pattern Recognition Conference}, 5360--5370.

\end{thebibliography}
	\end{spacing}
}

% TRYING OUT WITH APPENDIX
\clearpage
\onecolumn  
\appendix

% Unnumbered appendix heading:
\section*{Appendix A}

% Reset figure counter and define appendix-style figure numbers:
\setcounter{figure}{0}
\renewcommand{\thefigure}{A.\arabic{figure}}
\newpage
\begin{figure*}[h]
  \centering
  \includegraphics[width=0.6\textwidth]{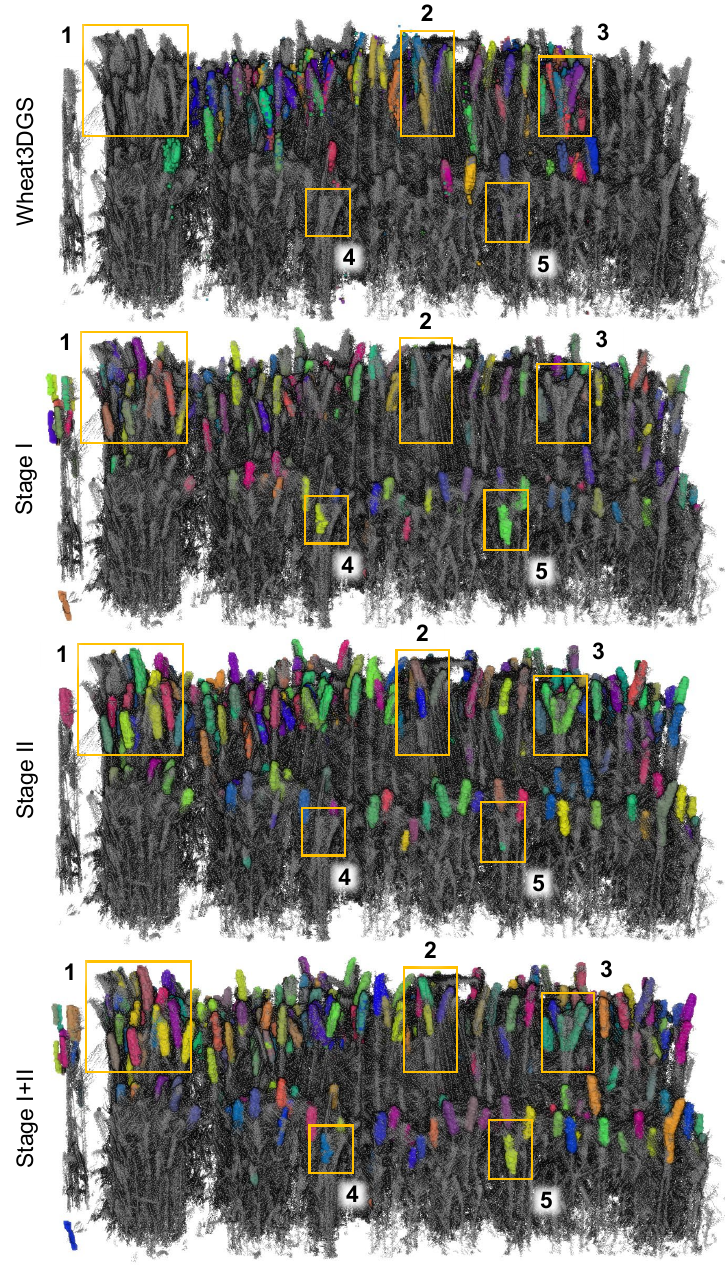}
  \caption{Additional side-view of the detected 3D wheat head instances (random color per instance alignment) overlaid over $P_t$ point cloud of the hold-out (test) plot; orange rectangles (1-5) marking regions with visually perceivable differences in instance segmentation performance between the realizations.}
  \label{fig:a1}
\end{figure*}

\end{document}